\documentclass[10pt,twocolumn,letterpaper]{article}

\usepackage{multirow}
\usepackage{color}
\usepackage{xcolor}
\usepackage{colortbl}
\usepackage{booktabs}

\usepackage{diagbox}

\usepackage{marvosym}
\makeatletter
\def\@fnsymbol#1{\ensuremath{\ifcase#1\or \dagger\or \ddagger\or
   \mathsection\or \mathparagraph\or \|\or **\or \dagger\dagger
   \or \ddagger\ddagger \else\@ctrerr\fi}}
\makeatother

\usepackage{loopsight}      %

\definecolor{linkblue}{rgb}{0.21,0.49,0.74}
\usepackage[pagebackref,breaklinks,colorlinks,allcolors=linkblue]{hyperref}

\newcommand{\rblue}{\rowcolor{blue!10}}

\title{LoopSight: Looped Vision Transformers with Gated Prediction Memory}

\author{
Zhaozhi Wang\textsuperscript{$1$}, Mingrui Wu\textsuperscript{$2$}, Jiaming Zhao\textsuperscript{$2$}, An-Syu Li\textsuperscript{$1$}, Jun Luo\textsuperscript{$1$}\thanks{Correspondence to \Letter \ junluo@ntu.edu.sg.}
\\
\\
\textsuperscript{$1$}Nanyang Technological University \quad \textsuperscript{$2$}University of Chinese Academy of Sciences}

\begin{document}
\maketitle
\begin{abstract}
Looped vision transformers extend computation for visual reasoning by repeatedly applying a shared Transformer core. A challenge is to preserve useful prediction information across loop iterations while allowing the computation state to evolve. In this paper we introduce LoopSight, a looped vision transformer that uses Gated Prediction Memory (GPM) to maintain a persistent prediction state for readout, separate from the core's evolving computation state. At each iteration, a cross-attention gate uses the core's candidate features, the previous prediction state, and the current task state to control writes at the patch and channel level. The model decodes its prediction from the updated memory. We condition the shared core on the current iteration using asymptotic loop encoding and train the model by supervising its prediction at loop counts sampled from a truncated Gaussian distribution. LoopSight outperforms LoopViT across three model scales on ARC-AGI-1 and ARC-AGI-2. On ARC-AGI-1, LoopSight-Small achieves 67.4\% pass@2, a gain of 6.6 percentage points over LoopViT-Small with comparable parameter counts and FLOPs. LoopSight also outperforms LoopViT on Visual Sudoku and ImageNet classification. Further analyses on ARC-AGI-1 show that LoopSight preserves more already-correct predictions and maintains more stable readout features as its computation state continues to evolve. The code will be released at {\color{magenta}https://github.com/feufhd/LoopSight}.
\end{abstract}    
\section{Introduction}
\label{sec:intro}

Studying visual reasoning helps us understand how models identify relevant structure in visual inputs and use it to generalize from limited experience. The Abstraction and Reasoning Corpus (ARC) tests this ability through grid transformation tasks that require a model to infer a rule from a few input--output examples and apply it to a new grid~\citep{chollet2019arc}. VARC formulates these tasks as task-conditioned image-to-image prediction on two-dimensional canvases~\citep{hu2026varc}. Inspired by prior work on recurrent computation in language modeling~\citep{dehghani2019universal,geiping2025huginn}, LoopViT extends this visual formulation by repeatedly applying a shared Transformer core to its hidden state, increasing computational depth at a fixed parameter count~\citep{shu2026loopvit}. This design may offer a new paradigm for parameter-efficient visual reasoning.

For looped models, a key challenge at inference time is that different inputs may benefit from different numbers of iterations~\citep{zhu2025ouro}. In our preliminary observations, a LoopViT prediction that matched the target at an intermediate iteration became incorrect after further updates (Fig.~\ref{fig:teaser}(b)). This raises the question of how to preserve useful predictive information while allowing further refinement. One possible approach is to introduce a separate readout branch with its own prediction state. By retaining useful features from earlier iterations and selectively incorporating new information from later ones, this branch could make the readout more stable as recurrent computation continues.

Following this principle, we introduce \textbf{LoopSight}, which extends LoopViT with a persistent readout state separate from the evolving computation state (Fig.~\ref{fig:teaser}(a)). To retain useful prediction features across loop iterations, LoopSight maintains the readout state through a dedicated branch called Gated Prediction Memory (GPM). At each loop iteration, GPM uses a cross-attention write gate to control how much of the core's new prediction features to write, using the current task state and the previous readout state as context. LoopSight decodes the final prediction from the updated readout state.

To help LoopSight adapt to different loop counts, we combine asymptotic loop encoding with variable-depth training. The encoding gives the shared core a conditioning signal that depends on the current loop iteration and approaches a learned limit as computation proceeds. During training, we sample the loop count from a truncated Gaussian distribution centered on a reference budget and supervise the prediction at the sampled depth.

Our contributions can be summarized as follows:
\begin{itemize}
    \item We introduce LoopSight, a looped vision transformer that maintains a persistent readout state to support reliable predictions across loop iterations, separate from the core's evolving computation state.
    \item To retain useful prediction features while incorporating new ones, we design Gated Prediction Memory (GPM). Its cross-attention write gate uses the current task state and the previous iteration's readout state to control how much each new prediction feature contributes to the updated readout state.
    \item To help LoopSight adapt to different loop counts, we combine asymptotic loop encoding with variable-depth training that samples loop counts from a truncated Gaussian distribution.
\end{itemize}

We evaluate LoopSight across three model scales on ARC-AGI-1 and ARC-AGI-2 and demonstrate consistent improvements over LoopViT. For example, LoopSight-Small achieves \textbf{67.4\%} pass@2 accuracy on ARC-AGI-1, surpassing LoopViT-Small by \textbf{+6.6} percentage points with comparable parameter counts and FLOPs. LoopSight also outperforms LoopViT on Visual Sudoku and ImageNet-100 classification. Prediction-trajectory and feature-evolution analyses on ARC-AGI-1 further show improved readout stability across recurrent iterations while the recurrent computation state continues to evolve.

\section{Related Work}
\label{sec:related-work}

\subsection{Visual Abstract Reasoning}

Visual abstract reasoning tests whether models can infer a rule from a few examples and apply it to a new visual configuration. Procedurally Generated Matrices (PGM)~\cite{barrett2018pgm} and RAVEN~\cite{zhang2019raven} probe relational generalization through matrix completion, and Bongard-LOGO~\cite{nie2020bongardlogo} tests whether models can distinguish visual concepts from a few examples. ARC~\cite{chollet2019arc} couples rule induction with execution by requiring an output grid with the correct dimensions and cell colors. ARC-AGI-2~\cite{chollet2025arcagi2} retains this format and introduces a set of grid transformations designed to challenge current reasoning systems. In this setting, each inferred rule is tested by the spatial transformation it produces on a held-out input.

In visual recognition, Feature Pyramid Networks (FPN)~\cite{fpn} fuse features across spatial scales, and iTPN~\cite{itpn} jointly pre-trains a visual backbone and its feature pyramid. For ARC, test-time training on augmented demonstrations helps language models adapt to new tasks~\cite{akyurek2025ttt}. VARC~\cite{hu2026varc} combines task adaptation with a visual formulation of ARC as image-to-image prediction, and neural cellular automata~\cite{etcheverry2026nca} solve grid tasks through local recurrent updates. Working directly with grids gives visual solvers access to the spatial relations needed both to infer a rule and to apply it through successive updates.

\subsection{Looped Visual Reasoning}

Reusing a Transformer block allows models to refine latent states over multiple steps at a fixed parameter count. Universal Transformers~\cite{dehghani2019universal} use shared self-attention for iterative refinement, and Huginn~\cite{geiping2025huginn} scales inference computation through latent recurrence. Constructive results also show that a looped Transformer with explicitly programmed weights can execute iterative algorithms~\cite{giannou2023looped}. In ARC and other puzzle tasks, HRM~\cite{wang2025hrm} organizes recurrence across two timescales, and TRM~\cite{jolicoeur2025trm} refines a candidate answer alongside a latent reasoning state. In visual ARC, this iterative formulation can be applied directly to spatial representations of the grid.

LoopViT~\cite{shu2026loopvit} performs ARC grid transformations by repeatedly applying a shared visual Transformer core. TraceViT~\cite{liu2026tracevit} extends this approach by supervising intermediate predictions with ordered transformation traces. Loop-OWM~\cite{gao2026loopowm} updates the grid state through spatial propagation and object-conditioned corrections. When the output grid is decoded from the current recurrent state, updates that fix errors in one region can also change predictions in regions that were already correct. The stability of the output then depends on which features each update preserves.

\begin{figure*}[ht]
    \centering
    \includegraphics[width=\linewidth]{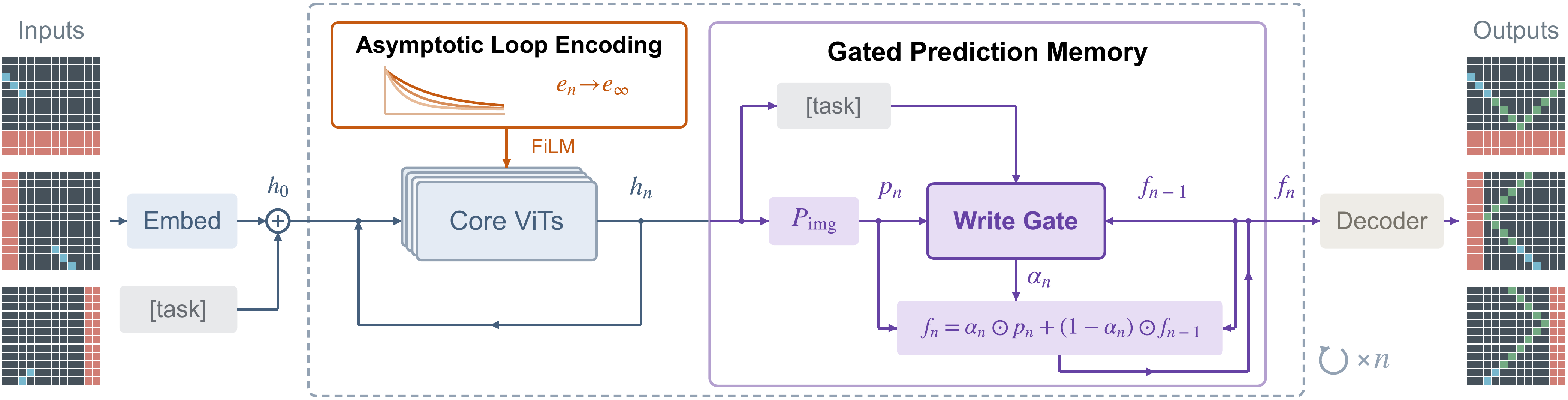}
    \caption{\textbf{Overview of LoopSight.} A shared ViT core iteratively updates the hidden state, conditioned by asymptotic loop encodings through feature-wise linear modulation (FiLM). At each iteration, a task-conditioned write gate controls how the current candidate updates the prediction memory. The final memory is decoded into the output grid.}
    \label{fig:framework}
\end{figure*}

\subsection{Computational Depth and Stability}

The stopping point in recurrent inference determines both the amount of computation and the state used for prediction. Adaptive Computation Time~\cite{graves2016act} and PonderNet~\cite{banino2021pondernet} learn when to stop recurrent computation, and Ouro~\cite{zhu2025ouro} uses learned exit gates to adapt recurrent depth to the input in language modeling. For ARC, USRL~\cite{chen2026usrl} uses the consistency of the current candidate solution with the demonstration examples as a stopping criterion. In image classification, MSDNet~\cite{huang2018msdnet} places classifiers at multiple depths to support prediction under different budgets. LoopFormer~\cite{jeddi2026loopformer} aligns short and full trajectories to support inference budgets up to its maximum training depth. On algorithmic tasks, sampling loop counts during training can reduce sensitivity to inference depth~\cite{kuo2026stochastic}. The benefit of a larger inference budget depends on how predictions evolve across the additional iterations.

Architectural choices also affect how signals propagate as computational depth increases. ResNet~\cite{resnet} uses identity shortcuts, and mHC~\cite{mhc} constrains mixing between residual streams to stabilize propagation across layers. In recurrent models, additional steps can help solve harder algorithmic instances~\cite{schwarzschild2021algorithm}, and work on equilibrium models and path independence~\cite{bai2019deq,anil2022path} studies convergence under repeated updates. Empirical studies have also documented overthinking on algorithmic tasks~\cite{bansal2022overthinking,zhou2026retain}. In visual ARC, this can manifest as a grid prediction that matches the target at one iteration and becomes incorrect after further updates. This motivates studying whether additional iterations can correct remaining errors while preserving predictions that are already correct.

\section{Method}
\label{sec:method}

To improve prediction stability during iterative visual reasoning, we propose \textbf{LoopSight}, which separates the evolving computation state from a persistent prediction state (Fig.~\ref{fig:framework}). Its three components address the retention of prediction information, iteration-dependent conditioning, and supervision at different stopping points:
\begin{itemize}
    \item \textbf{Gated Prediction Memory (GPM)} retains prediction features across iterations and selectively incorporates new information through a task-conditioned write gate.
    \item \textbf{Asymptotic Loop Encoding} supplies iteration-dependent conditioning that approaches a learned limiting representation as computation continues.
    \item \textbf{Training Strategy} exposes the prediction memory to supervision at multiple stopping points by sampling loop counts from a truncated Gaussian distribution and supervising the final prediction at each sampled endpoint.
\end{itemize}

\subsection{Gated Prediction Memory}
\label{sec:gpm}

Continued recurrent computation may alter information that already supports a correct prediction. To retain useful prediction information while allowing further refinement, we introduce \textbf{Gated Prediction Memory (GPM)}, a separate prediction-feature state maintained across iterations alongside the main computational stream.

For a forward pass of $N$ iterations, let $n\in\{1,\ldots,N\}$ denote the iteration index.
Following LoopViT~\citep{shu2026loopvit}, we embed the input grid into patch tokens and prepend a task-specific token to construct the initial hidden state $h_0$. At iteration $n$, the weight-shared Transformer core updates $h_{n-1}$ conditioned on an iteration encoding $e_n$, followed by RMSNorm to obtain $h_n$. The core adopts a hybrid Transformer architecture, combining self-attention with convolutional gated feed-forward networks.

\paragraph{Gated Memory Update.}
At each iteration, a shared linear projection $P_{\mathrm{img}}$ first maps the image tokens of $h_n$ to a prediction candidate $p_n$. 
We maintain a persistent memory $f_n$ with the same dimensions as $p_n$, updated through a gated interpolation:
\begin{equation}
    f_n = \alpha_n \odot p_n
        + (1-\alpha_n)\odot f_{n-1},
    \label{eq:memory-update}
\end{equation}
where $\alpha_n$ contains element-wise write coefficients in $[0,1]$. Small coefficients favor retaining the previous memory, while larger coefficients favor the current candidate. Thus, $f_n$ selectively integrates new prediction features with previously retained information, separately from the main recurrent computation.
We initialize $f_0=0$ and set $\alpha_1=1$, yielding $f_1=p_1$. After the final iteration, $f_N$ is decoded into pixel-wise
class logits through layer normalization and a linear prediction head.

\begin{figure}[t]
    \centering
    \includegraphics[width=\linewidth]{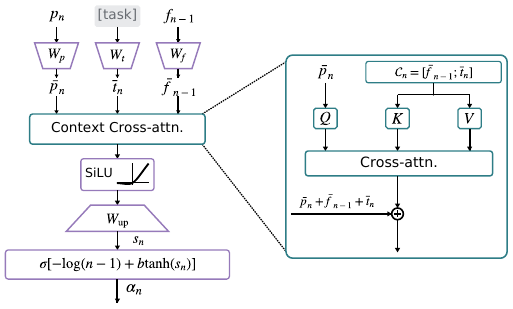}
    \caption{\textbf{Write gate architecture.} Separate down-projections produce candidate, memory, and task features. Candidates query memory and task context through cross-attention, followed by a local residual sum, SiLU~\citep{silu}, and $W_{\mathrm{up}}$ to generate gate scores. The sigmoid applies for $n>1$; the first write uses $\alpha_1=1$.}
    \label{fig:write-gate}
\end{figure}

\paragraph{Task-conditioned Write Gate.}
To determine the write coefficients, we first compute content-dependent write scores $s_n$ from the current prediction candidate $p_n$, previous memory $f_{n-1}$, and task state $t_n$ extracted from the task token in $h_n$. To incorporate global context from the prediction memory, we employ a cross-attention gate (Fig.~\ref{fig:write-gate}).

We apply RMSNorm and separate linear projections $W_p$, $W_f$, and $W_t$ to obtain low-dimensional features $\bar p_n$, $\bar f_{n-1}$, and $\bar t_n$, respectively. The context sequence $\mathcal{C}_n=[\bar f_{n-1};\bar t_n]$ combines previous memory features with the current task state.
Using $\bar p_n$ as queries and $\mathcal{C}_n$ as keys
and values, the gate computes:
\begin{equation}
\begin{aligned}
    s_n = W_{\mathrm{up}}\Bigl(\operatorname{SiLU}\bigl(
    &\bar p_n+\bar f_{n-1}+\bar t_n \\
    &+\operatorname{CrossAttn}(\bar p_n,\mathcal{C}_n)
    \bigr)\Bigr),
\end{aligned}
\label{eq:write-score}
\end{equation}
where $\bar t_n$ is broadcast across patches and
$W_{\mathrm{up}}$ projects the features back to the
original channel dimension. The resulting content-dependent scores $s_n$ are then used to determine the write coefficients $\alpha_n$.

\paragraph{Loop-calibrated Updates.}
To moderate updates to accumulated prediction features while retaining content-dependent revision, we calibrate the write scores $s_n$ with an inverse-iteration schedule:
\begin{equation}
    \alpha_n = \operatorname{sigmoid}\!
    \left[-\log(n-1)+b\tanh(s_n)\right],
    \qquad n>1,
    \label{eq:write-gate}
\end{equation}
where $b>0$ is a fixed hyperparameter controlling the maximum adjustment.
At initialization, the weights and bias of $W_{\mathrm{up}}$ are zero, so $s_n=0$ and $\alpha_n=1/n$. The memory therefore initially implements uniform averaging of the prediction candidates: $f_n=\frac{1}{n}\sum_{i=1}^{n}p_i$.
Since $\tanh(s_n)\in[-1,1]$, the write coefficients are bounded by
\begin{equation}
    \frac{1}{1+\exp(b)(n-1)}
    \leq \alpha_n \leq
    \frac{1}{1+\exp(-b)(n-1)}.
    \label{eq:gate-bounds}
\end{equation}
For fixed $b$, the bounds give $\alpha_n=\Theta(1/n)$ and progressively cap each candidate’s weight. Content-dependent weighting within these bounds allows selective updates and favors gradual refinement of the prediction memory.

\subsection{Asymptotic Loop Encoding}
\label{sec:asymptotic-encoding}

To make recurrent updates aware of their stage of computation while moderating changes in the conditioning signal at later iterations, we introduce \textbf{Asymptotic Loop Encoding}. Its iteration-dependent component decays toward zero, yielding a conditioning signal that approaches a learned limiting representation.
Let $a_n$ be a vector of geometric features at iteration $n$. Each channel, indexed by $k$, has its own learned positive decay rate $\lambda_k$ and is defined as
\begin{equation}
    a_n^{(k)}=(1-e^{-\lambda_k})e^{-\lambda_k(n-1)},
    \qquad \lambda_k>0,
    \label{eq:geometric-encoding}
\end{equation}
where $a_n^{(k)}$ denotes the $k$-th entry of $a_n$ and the scalar exponential terms define a geometric decay over iterations. We obtain each $\lambda_k$ by applying softplus to an unconstrained learned scalar, ensuring positivity.
The factor $1-e^{-\lambda_k}$ makes each channel sum to one over positive iterations, and every channel approaches zero as $n$ grows. Larger decay rates emphasize early iterations, while smaller rates retain variation over longer ranges. These features are projected into the hidden dimension to form the loop encoding:
\begin{equation}
    e_n=e_{\infty}+W_e a_n,
    \qquad \lim_{n\to\infty}e_n=e_{\infty},
    \label{eq:loop-encoding}
\end{equation}
where $W_e$ is a learned linear projection and $e_{\infty}$ is a learned limiting embedding. The difference between consecutive encodings tends to zero, moderating changes in conditioning at later iterations. Since $e_n$ depends only on the absolute iteration index $n$, the same iteration receives the same conditioning regardless of the selected stopping point $N$.

To incorporate the loop encoding into the shared Transformer core, we employ feature-wise linear modulation (FiLM)~\citep{perez2018film}.
For each block $j$, we add a projected cosine encoding of its relative block position to $e_n$ and denote the resulting conditioning vector by $c_{n,j}$.
A block-specific two-layer MLP then maps $c_{n,j}$ to separate scale and shift parameters $(\gamma,\beta)$ for the attention and convolutional feed-forward branches.
For either branch, the modulation is applied as
\begin{equation}
    \operatorname{FiLM}(z)
    =(1+\gamma)\odot\operatorname{RMSNorm}(z)+\beta,
    \label{eq:film}
\end{equation}
where $z$ is the token sequence entering the branch and $\operatorname{RMSNorm}(z)$ is its token-wise normalized representation. The final layer of each modulation MLP is zero-initialized, yielding $\gamma=\beta=0$ at initialization. The modulation MLPs share parameters across iterations, providing a consistent conditioning mechanism throughout recurrent computation.

\subsection{Training Strategy}
\label{sec:training-strategy}

\paragraph{Variable-depth Training.}
Supervision at a single fixed endpoint does not directly optimize prediction quality at other loop steps. To expose the prediction memory to supervision at different stopping points, we adopt variable-depth training by sampling the number of iterations $N$ from a truncated discrete Gaussian distribution:
\begin{equation}
    \Pr(N)\propto
    \exp\!\left[-\frac{(N-\mu)^2}{2\tau^2}\right],
    \quad N\in\{N_{\min},\ldots,N_{\max}\},
    \label{eq:depth-sampling}
\end{equation}
where $\Pr(N)$ is the probability of sampling $N$ loop steps, $N_{\min}$ and $N_{\max}$ are the minimum and maximum allowed depths, and $\mu$ and $\tau>0$ are the center and scale of the underlying Gaussian. This sampling strategy emphasizes depths near $\mu$ while also covering shorter and longer executions. At each optimizer update, one depth is sampled and shared across all accumulated microbatches.

After $N$ iterations, we supervise the output decoded from the final prediction memory $f_N$ using pixel-wise cross-entropy:
\begin{equation}
    \mathcal{L}=\mathbb{E}_{N}\!\left[
        \operatorname{CE}\bigl(\operatorname{Dec}(f_N),y\bigr)
    \right],
    \label{eq:training-loss}
\end{equation}
where $\mathcal{L}$ is the training objective, $\mathbb{E}_N$ denotes expectation over the sampled depths, $\operatorname{Dec}$ is the shared decoder, and $y$ denotes the target grids. Gradients propagate through all unrolled iterations, jointly optimizing the shared core, prediction memory, and loop encoding. Supervising different endpoints encourages useful readouts at multiple stages of recurrent computation and complements GPM's retention and revision mechanism.

\begin{table*}[t]
\centering
\small
\setlength{\tabcolsep}{4pt}
\caption{ARC-AGI results (pass@2, \%) by model scale. Bold denotes the best score in each scale group and benchmark. Parentheses report gains over the strongest baseline in the group. $^{\S}$HRM/TRM Loop Steps denote outer steps $\times$ module calls per outer step.}
\label{tab:arc-comparison}
\begin{tabular}{l<{\hspace{8pt}}r<{\hspace{6pt}}r<{\hspace{4pt}}r<{\hspace{8pt}}r<{\hspace{6pt}}r}
\toprule
\multirow{2}{*}{Model} & \multirow{2}{*}{Loop Steps} & \multirow{2}{*}{Params} & \multirow{2}{*}{FLOPs} & \multirow{2}{*}{ARC-AGI-1} & ARC-AGI-2 \\
 & & & & & (ARC-AGI-1 pre-training) \\
\midrule
\multicolumn{6}{l}{\textit{Small (7M$-$)}} \\
\addlinespace[2pt]
TRM$^{\S}$ & $16\times 15$ & 6.8M & 3.8T & 44.6 & - \\
NCA & - & 3.5M & 735G & 48.8 & - \\
LoopViT-Small & 20--28 & 3.8M & 260--364G & 60.1 & 10.0 \\
LoopViT-Small (reproduced) & 12 & 3.8M & 156G & 60.8 & 10.0 \\
\rblue \textbf{LoopSight-Small} & 12 & 4.3M & 158G & \textbf{67.4 (+6.6)} & \textbf{13.2 (+3.2)} \\
\midrule
\multicolumn{6}{l}{\textit{Medium (10--12M)}} \\
\addlinespace[2pt]
NCA (ensemble) & - & 11M & 2.2T & 51.3 & - \\
LoopViT-Medium & 4--8 & 11M & 132--261G & 63.8 & 11.5 \\
\rblue \textbf{LoopSight-Medium} & 6 & 10M & 178G & \textbf{67.5 (+3.7)} & \textbf{15.4 (+3.9)} \\
\midrule
\multicolumn{6}{l}{\textit{Large (18M+)}} \\
\addlinespace[2pt]
HRM$^{\S}$ & $16\times 6$ & 27M & 3.0T & 40.3 & - \\
VARC-ViT & - & 18M & 56G & 54.5 & 8.3 \\
VARC (ensemble) & - & 73M & 230G & 60.4 & 11.1 \\
LoopViT-Large & 4--8 & 18M & 218--433G & 65.8 & 14.2 \\
\rblue \textbf{LoopSight-Large} & 6 & 19M & 329G & \textbf{69.6 (+3.8)} & \textbf{16.0 (+1.8)} \\
\bottomrule
\end{tabular}
\end{table*}

\paragraph{Task Adaptation and Inference.}
Following LoopViT, we adapt the model to each unseen task by initializing new task embeddings and fine-tuning all model parameters on augmented demonstration pairs, using the same final-output objective with a fixed number of iterations. At inference, we select an iteration budget $N$ and decode predictions from the final memory $f_N$. The predictions are mapped back by inverting the augmentations, and complete candidate grids are ranked by their frequency across predictions.

\section{Experiment}
\label{sec:experiment}

We evaluate LoopSight on ARC-AGI-1 and ARC-AGI-2 across three model scales. 
Analyses of prediction trajectories, readout features, and memory writes examine how prediction information is retained as the recurrent core evolves. Additional experiments compare alternative readout designs and use progressive ablations to quantify the contributions of GPM, asymptotic loop encoding, and variable-depth training. Finally, the evaluation extends beyond ARC to grid completion on Visual Sudoku and image classification on ImageNet-100.

\subsection{Experimental Setup}

We pretrain on the 400 ARC-AGI-1 training tasks and synthetic RE-ARC examples~\citep{rearc}, then evaluate on the public ARC-AGI-1 and ARC-AGI-2 sets of 400 and 120 tasks, respectively. For each task, we aggregate grid votes from two test-time training (TTT) runs with ten inference augmentations each. Pass@2 requires an exact match among the two most-voted distinct grids. We average query scores within each task and then across tasks.

We evaluate Small, Medium, and Large LoopSight models. Offline training samples loop counts from 6--18 for Small and 3--9 for Medium and Large, with supervision at the sampled stopping point. Unless otherwise specified, TTT and inference use 12 loops for Small and 6 for Medium and Large. Please refer to Sec.~\ref{app:experimental-settings} in Appendix for implementation details.

\subsection{Main Results}

LoopSight consistently outperforms existing methods across all three model scales on ARC-AGI-1, achieving 67.4\% (+6.6), 67.5\% (+3.7), and 69.6\% (+3.8) pass@2 for Small, Medium, and Large, respectively, with gains measured against the strongest baseline in each group, Table~\ref{tab:arc-comparison}. 
Notably, LoopSight-Small surpasses LoopViT-Large (65.8\%) with substantially fewer parameters and FLOPs.

Following pretraining on ARC-AGI-1 and task-specific adaptation on ARC-AGI-2, LoopSight achieves 13.2\% (+3.2), 15.4\% (+3.9), and 16.0\% (+1.8) pass@2 across the three scales, outperforming the strongest baselines in each group, respectively. LoopSight-Medium also surpasses LoopViT-Large (14.2\%) with fewer parameters and FLOPs.
Within our LoopSight variants, performance consistently improves with model scale on both benchmarks.

\begin{figure}[h]
    \centering
    \includegraphics[width=0.99\columnwidth]{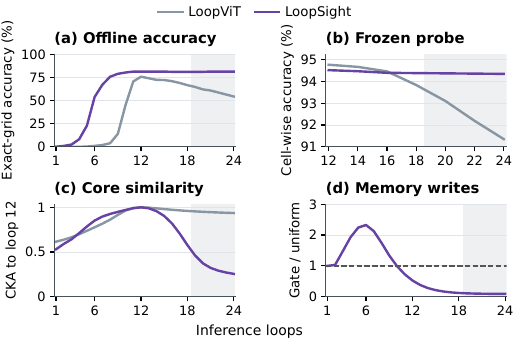}
    \vspace{-2mm}
    \caption{\textbf{Prediction and readout stability across loops.}
    (a) Offline exact-grid accuracy of Small checkpoints on 416 test queries from the 400 ARC-AGI-1 training tasks.
    (b) Cell-wise accuracy of linear probes fitted to each model's loop-12 readout features and held fixed across loops, averaged over 32 probe test tasks.
    (c) Spatial centered kernel alignment (CKA) of recurrent core features at loops 1--24 relative to loop 12, averaged over 128 tasks.
    (d) LoopSight's mean GPM write coefficient at loop $n$ relative to the uniform accumulation baseline $1/n$, averaged over the same 128 tasks. The horizontal dashed line marks the baseline.
    Gray shading marks inference depths beyond LoopSight-Small's training maximum of 18 loops.}
    \label{fig:loop-step-analysis}
\end{figure}

\subsection{Prediction Stability Across Loops}
\label{sec:loop-step-generalization}

To investigate whether LoopSight preserves correct predictions as the loop count increases, we evaluate frozen LoopViT/LoopSight-Small checkpoints at every depth from 1 to 24 loops.

\begin{figure*}[t]
    \centering
    \includegraphics[width=\textwidth]{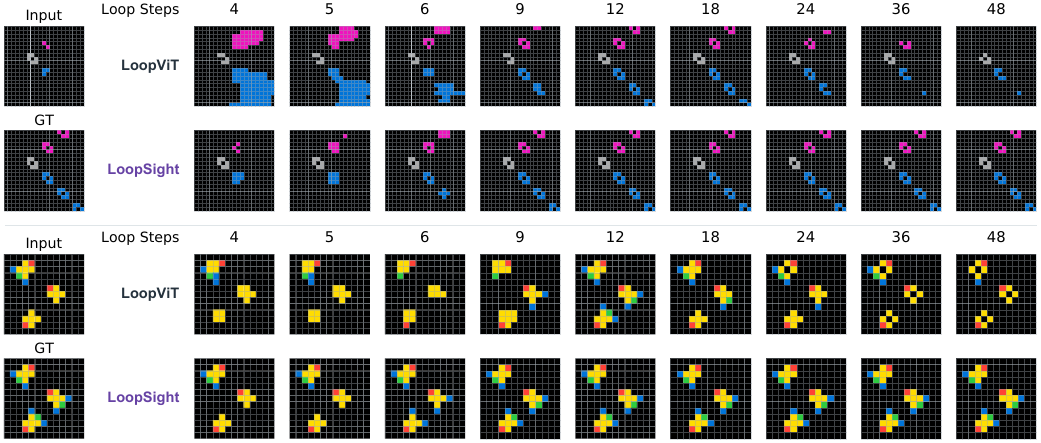}
    \vspace{-5.5mm}
    \caption{\textbf{Prediction
    trajectories across loop steps.}
    Illustrative training and held-out examples from ARC-AGI-1, with predictions cropped to the ground-truth (GT) extent.}
    \label{fig:loop-step-examples}
\end{figure*}

Fig.~\ref{fig:loop-step-analysis}(a) shows that LoopViT's exact-grid accuracy peaks at loop 12 and declines with further looping, while LoopSight achieves a higher peak accuracy than LoopViT and stays near that peak as the loop count increases to 24. Fig.~\ref{fig:loop-step-examples} illustrates how predictions evolve across loops, showing that LoopViT can produce a correct grid and then lose it with further looping. LoopSight preserves its correct predictions across the later loops shown. Together, these results suggest that LoopSight's readout can preserve correct grid predictions as the loop count increases, even beyond its training range.

We assess readout stability by fitting a linear probe to each model's loop-12 readout features to predict the output grid and holding the probe fixed at later loops (Fig.~\ref{fig:loop-step-analysis}(b)). To track core evolution, we compute linear centered kernel alignment (CKA)~\citep{cka} on matched patches from 128 tasks, comparing recurrent core features at loops 1--24 with those at loop 12 (Fig.~\ref{fig:loop-step-analysis}(c)). From loop 12 to loop 24, LoopViT's core CKA remains high and its frozen probe's cell-wise accuracy drops from 94.8\% to 91.4\%. Over the same interval, LoopSight's core CKA drops substantially and its frozen-probe accuracy stays nearly constant (94.5\% to 94.3\%). Beyond its 12-loop training depth, LoopViT repeatedly applies the shared core with the same final step embedding. In this regime, successive updates are highly aligned and become smaller relative to the spatially centered core state, leaving the overall patch geometry largely intact. LoopViT reads directly from the current state, so drift in task-relevant features can still disrupt earlier predictions. LoopSight's GPM instead uses a learned gate to control how much each new candidate contributes to a separate prediction memory (Eq.~\ref{eq:memory-update}), allowing the core to evolve further while retaining the readout that supports earlier predictions.

To see how GPM stabilizes the readout, we compare its write coefficients with the uniform baseline $1/n$ (Fig.~\ref{fig:loop-step-analysis}(d)). The coefficients rise above the baseline early and fall well below it at later loops. Stronger early writes incorporate new features as the prediction forms, and weaker later writes limit changes to the accumulated readout as the core continues to evolve.

\subsection{Readout Ablation}
\label{sec:readout-ablation}

To explore how different readout designs perform, we compare five readout variants on ARC-AGI-1 using the same offline pretraining, test-time training, and inference settings, and the results are reported in Table~\ref{tab:readout-ablation}.

LoopViT decodes the final hidden state. Uniform fusion averages projected loop features, and learned-weight fusion applies softmax-normalized per-loop weights shared across examples. The scalar write gate computes an input-dependent write coefficient at each loop from pooled summaries of the candidate and previous memory. That coefficient applies to all patches and channels of the example. GPM assigns a coefficient to each patch and channel using cross-attention conditioned on the current candidate, previous memory, and task state.

Uniform fusion raises pass@2 from 60.8\% to 61.8\%, suggesting that projected features from earlier loops can complement the final hidden state. Learned-weight fusion further increases pass@2 to 62.3\%, a gain of 0.5 percentage points over uniform fusion. The scalar-gated readout gains another +2.1, consistent with a benefit from adjusting the contribution of a loop to the input and the information already accumulated. 
Furthermore, LoopSight with GPM achieves 67.4\% pass@2, a gain of 3.0 percentage points over the scalar-gated readout, supporting the effectiveness of adaptively fusing features across loop steps at the patch and channel level.

\begin{table}[t]
    \centering
    \caption{Readout ablation on ARC-AGI-1.}
    \label{tab:readout-ablation}
    \small
    \begin{tabular*}{0.85\columnwidth}{@{\extracolsep{\fill}}lr@{}}
        \toprule
        Readout & Pass@2 (\%) \\
        \midrule
        LoopViT (final state) & 60.8 \\
        Uniform fusion & 61.8 \\
        Learned-weight fusion & 62.3 \\
        Scalar write gate & 64.4 \\
        \textbf{LoopSight (GPM)} & \textbf{67.4} \\
        \bottomrule
    \end{tabular*}
    \vspace{-5mm}
\end{table}

\subsection{Progressive Ablation Studies}
\label{sec:ablation}

To assess the incremental contribution of each component, we add GPM, asymptotic loop encoding (ALE), and variable-depth training sequentially to LoopViT. We use a compact backbone with three shared Transformer blocks, hidden dimension 192, three attention heads, and feed-forward dimension 512. All configurations use the same fixed subsets of 192 ARC-AGI-1 training tasks with RE-ARC augmentation and 192 evaluation tasks. We train for 50 offline epochs and use two 50-epoch TTT attempts per task at a fixed depth of six loops. We vary only the inference depth across columns in Table~\ref{tab:ablation}.

The first three rows use a fixed offline depth of six loops. Adding GPM retains the learned additive loop embeddings, and adding ALE replaces them with FiLM-based loop conditioning. For the final row, we sample the offline depth from 3--9 for each batch using a truncated Gaussian with center 6 and scale 1.5, and supervise the final prediction at that sampled depth. The symmetric sampling distribution has an expected depth of six, so the final comparison varies the supervised stopping point at the same expected offline loop count and fixed TTT depth.

\begin{table}[t]
    \centering
    \caption{Progressive ablation on ARC-AGI-1. Entries report pass@2 (\%) at inference depth $N$. Each addition builds on the preceding row.}
    \label{tab:ablation}
    \small
    \setlength{\tabcolsep}{2pt}
    \begin{tabular*}{\columnwidth}{@{\extracolsep{\fill}}lcccc@{}}
        \toprule
        Configuration & $N{=}3$ & $N{=}6$ & $N{=}9$ & $N{=}12$ \\
        \midrule
        LoopViT & 15.6 & 42.7 & 41.7 & 38.0 \\
        + GPM & 17.2 & 45.3 & 43.8 & 40.1 \\
        + ALE & 17.2 & 45.8 & 44.8 & 42.7 \\
        + Gaussian & \multirow{2}{*}{\textbf{28.1}} & \multirow{2}{*}{\textbf{46.9}} & \multirow{2}{*}{\textbf{46.9}} & \multirow{2}{*}{\textbf{46.4}} \\
        \textbf{(LoopSight)} & & & & \\
        \bottomrule
    \end{tabular*}
\end{table}

Adding GPM improves pass@2 at every evaluated stopping point, with gains from +1.6 to +2.6 under fixed-depth supervision (Table~\ref{tab:ablation}). At the supervised depth of six loops, pass@2 rises from 42.7\% to 45.3\%. Adding ALE then yields +0.5 at six loops and larger gains of +1.0 and +2.6 at nine and twelve loops, respectively.

With truncated Gaussian sampling, the largest additional gain occurs at three loops, the earliest depth included in the supervision schedule. Pass@2 rises from 17.2\% to 28.1\% at this early readout, a gain of 10.9 percentage points, and reaches 46.9\% at six loops. Across the later stopping points at six, nine, and twelve loops, the spread in Pass@2 decreases from 3.1 to 0.5 percentage points compared with fixed-depth training with GPM and ALE. This narrower spread indicates that aggregate performance is less sensitive to the choice of stopping point across the tested later depths. At the original six-loop endpoint, Pass@2 also improves from 45.8\% to 46.9\%. This gain may stem from the regularizing effect of supervision at multiple stopping points, which encourages the model to produce reliable readouts across loop depths and thereby reduces specialization to a single endpoint.

\subsection{Beyond ARC-AGI}
\label{sec:task-generalization}

To explore whether LoopSight's effectiveness extend to tasks with different output structures, we evaluate grid completion on Visual Sudoku~\citep{etcheverry2026nca} and image-level classification on ImageNet-100~\citep{imagenet,cmc}.

\paragraph{Visual Sudoku.}
We train LoopViT and LoopSight from scratch on easy Sudoku puzzles rendered with handwritten digits, using matched settings and the same input resolution. Easy follows the training distribution, and Hard contains more challenging puzzles outside that distribution. We report whole-board accuracy, counting a board as correct when all cells match the solution. Qualitative examples are provided in Sec.~\ref{sec:qualitative-visual-sudoku} in Appendix. LoopSight reaches 89.6\% on Easy (+1.5) and 27.7\% on Hard (+2.7) over LoopViT at comparable parameter counts and FLOPs (Table~\ref{tab:visual-sudoku-pixel}). The larger gain occurs on puzzles outside the training distribution, where LoopViT completes 25.0\% of boards. LoopSight also exceeds the reported NCA accuracies on both splits at substantially lower FLOPs.

\begin{table}[t]
  \caption{Visual Sudoku whole-board accuracy (\%) and inference
  cost ($K=1$). The reported NCA results~\citep{etcheverry2026nca} use
  1,024 updates on Easy and 3,072 on Hard. The exact NCA Hard subset
  indices are unavailable.}
  \label{tab:visual-sudoku-pixel}
  \centering
  \small
  \setlength{\tabcolsep}{3.5pt}
  \renewcommand{\arraystretch}{1.05}
  \resizebox{0.94\columnwidth}{!}{%
  \begin{tabular}{@{}lcrrr@{}}
    \toprule
    Model & Params & FLOPs & Easy & Hard \\ \midrule
    NCA & 2.1M & 283.2T & 87.8 & -- \\
    NCA  & 2.1M & 849.5T & -- & 18.9 \\
    LoopViT & 4.4M & 43.3G & 88.1 & 25.0 \\ \rblue
    \textbf{LoopSight}
      & 5.1M & 44.0G
      & \textbf{89.6 (+1.5)} & \textbf{27.7 (+2.7)} \\
    \bottomrule
  \end{tabular}%
  }
\end{table}

\paragraph{ImageNet-100 Classification.}
We train the Small versions of LoopViT and LoopSight and a ViT-Small/16~\citep{vit} baseline from scratch for 100 epochs under a common optimization and augmentation protocol on ImageNet-100. All models use $224\times224$ inputs and $16\times16$ patches. For the recurrent models, we replace grid reconstruction with mean pooling, LayerNorm, and a linear classifier. LoopViT pools the final recurrent patch states, and LoopSight pools the final GPM states. LoopSight achieves 70.5\% top-1 accuracy, outperforming LoopViT by 3.0 percentage points with comparable parameter counts and FLOPs (Table~\ref{tab:imagenet100-classification}). 

\begin{table}[t]
  \caption{ImageNet-100 Top-1 accuracy (\%) and inference cost.
  LoopViT and LoopSight both use a hidden dimension of 256 and 6 Transformer blocks shared
  across 6 loops.}
  \label{tab:imagenet100-classification}
  \centering
  \small
  \setlength{\tabcolsep}{4pt}
  \renewcommand{\arraystretch}{1.05}
  \resizebox{0.74\columnwidth}{!}{%
  \begin{tabular}{@{}lrrr@{}}
    \toprule
    Model & Params & FLOPs & Top-1 Acc. \\ \midrule
    ViT-Small & 22M & 4.6G & 58.6 \\
    LoopViT & 3.5M & 4.5G & 67.5 \\ \rblue\textbf{LoopSight}
      & 4.0M & 4.6G & \textbf{70.5 (+3.0)}\\
    \bottomrule
  \end{tabular}%
  }
\end{table}

Together, these results suggest that LoopSight’s advantage over LoopViT extends across tasks with different output structures, from producing spatial predictions that jointly satisfy constraints to aggregating spatial evidence into a single image-level prediction.

\section{Conclusion}
\label{sec:conclusion}

We introduce LoopSight, a looped vision transformer that maintains a persistent prediction state separate from the core’s evolving computation state. Its Gated Prediction Memory branch uses task-conditioned writes at each step to selectively retain and update features in this state, supporting reliable predictions across loop iterations. Asymptotic loop encoding conditions the recurrent core on the current iteration, and variable-depth training supervises final predictions at multiple stopping points. LoopSight outperforms LoopViT at all three model scales on ARC-AGI-1 and ARC-AGI-2 and also improves accuracy on Visual Sudoku and ImageNet-100. Analyses on ARC-AGI-1 show that LoopSight preserves more already-correct predictions and maintains more stable readout features as its computation state continues to evolve. These findings point to the separation of computation and prediction as a design principle for looped vision transformers, enabling reliable readout during continued refinement.

{
    \small
    \bibliographystyle{ieeenat_fullname}
    \bibliography{main}

\begin{thebibliography}{49}
\providecommand{\natexlab}[1]{#1}
\providecommand{\url}[1]{\texttt{#1}}
\expandafter\ifx\csname urlstyle\endcsname\relax
  \providecommand{\doi}[1]{doi: #1}\else
  \providecommand{\doi}{doi: \begingroup \urlstyle{rm}\Url}\fi

\bibitem[Aky\"{u}rek et~al.(2025)Aky\"{u}rek, Damani, Zweiger, Qiu, Guo, Pari,
  Kim, and Andreas]{akyurek2025ttt}
Ekin Aky\"{u}rek, Mehul Damani, Adam Zweiger, Linlu Qiu, Han Guo, Jyothish
  Pari, Yoon Kim, and Jacob Andreas.
\newblock {The Surprising Effectiveness of Test-Time Training for Few-Shot
  Learning}.
\newblock In \emph{ICML}, 2025.

\bibitem[Anil et~al.(2022)Anil, Pokle, Liang, Treutlein, Wu, Bai, Kolter, and
  Grosse]{anil2022path}
Cem Anil, Ashwini Pokle, Kaiqu Liang, Johannes Treutlein, Yuhuai Wu, Shaojie
  Bai, J.~Zico Kolter, and Roger~B Grosse.
\newblock {Path Independent Equilibrium Models Can Better Exploit Test-Time
  Computation}.
\newblock In \emph{NeurIPS}, 2022.

\bibitem[Bai et~al.(2019)Bai, Kolter, and Koltun]{bai2019deq}
Shaojie Bai, J.~Zico Kolter, and Vladlen Koltun.
\newblock {Deep Equilibrium Models}.
\newblock In \emph{NeurIPS}, 2019.

\bibitem[Banino et~al.(2021)Banino, Balaguer, and
  Blundell]{banino2021pondernet}
Andrea Banino, Jan Balaguer, and Charles Blundell.
\newblock {PonderNet: Learning to Ponder}.
\newblock In \emph{8th ICML Workshop on Automated Machine Learning}, 2021.

\bibitem[Bansal et~al.(2022)Bansal, Schwarzschild, Borgnia, Emam, Huang,
  Goldblum, and Goldstein]{bansal2022overthinking}
Arpit Bansal, Avi Schwarzschild, Eitan Borgnia, Zeyad Emam, Furong Huang, Micah
  Goldblum, and Tom Goldstein.
\newblock {End-to-end Algorithm Synthesis with Recurrent Networks:
  Extrapolation without Overthinking}.
\newblock In \emph{NeurIPS}, 2022.

\bibitem[Barrett et~al.(2018)Barrett, Hill, Santoro, Morcos, and
  Lillicrap]{barrett2018pgm}
David Barrett, Felix Hill, Adam Santoro, Ari Morcos, and Timothy Lillicrap.
\newblock {Measuring abstract reasoning in neural networks}.
\newblock In \emph{ICML}, 2018.

\bibitem[Chen et~al.(2026)Chen, Li, Li, Zhuang, Wu, Li, Wang, Li, Hao, and
  Wu]{chen2026usrl}
Xinwang Chen, Xiuxing Li, Qing Li, Ziyue Zhuang, Yutong Wu, Ziyu Li, Zhuo Wang,
  Kai Li, Jianye Hao, and Xia Wu.
\newblock {Human-like Abstract Visual Reasoning via Understanding and Solving
  Reasoning Loop}.
\newblock In \emph{CVPR}, 2026.

\bibitem[Chollet(2019)]{chollet2019arc}
Fran{\c{c}}ois Chollet.
\newblock {On the Measure of Intelligence}.
\newblock arXiv preprint arXiv:1911.01547, 2019.

\bibitem[Chollet et~al.(2025)Chollet, Knoop, Kamradt, Landers, and
  Pinkard]{chollet2025arcagi2}
Fran{\c{c}}ois Chollet, Mike Knoop, Gregory Kamradt, Bryan Landers, and Henry
  Pinkard.
\newblock {ARC-AGI-2}: A new challenge for frontier {AI} reasoning systems.
\newblock \emph{arXiv preprint arXiv:2505.11831}, 2025.

\bibitem[Cubuk et~al.(2020)Cubuk, Zoph, Shlens, and Le]{randaug}
Ekin~Dogus Cubuk, Barret Zoph, Jon Shlens, and Quoc Le.
\newblock {RandAugment: Practical Automated Data Augmentation with a Reduced
  Search Space}.
\newblock In \emph{NeurIPS}, 2020.

\bibitem[Dehghani et~al.(2019)Dehghani, Gouws, Vinyals, Uszkoreit, and
  Kaiser]{dehghani2019universal}
Mostafa Dehghani, Stephan Gouws, Oriol Vinyals, Jakob Uszkoreit, and {\L}ukasz
  Kaiser.
\newblock {Universal Transformers}.
\newblock In \emph{ICLR}, 2019.

\bibitem[Deng et~al.(2009)Deng, Dong, Socher, Li, Li, and Fei-Fei]{imagenet}
Jia Deng, Wei Dong, Richard Socher, Li-Jia Li, Kai Li, and Li Fei-Fei.
\newblock {ImageNet: A large-scale hierarchical image database}.
\newblock In \emph{CVPR}, 2009.

\bibitem[Dosovitskiy et~al.(2021)Dosovitskiy, Beyer, Kolesnikov, Weissenborn,
  Zhai, Unterthiner, Dehghani, Minderer, Heigold, Gelly, Uszkoreit, and
  Houlsby]{vit}
Alexey Dosovitskiy, Lucas Beyer, Alexander Kolesnikov, Dirk Weissenborn,
  Xiaohua Zhai, Thomas Unterthiner, Mostafa Dehghani, Matthias Minderer, Georg
  Heigold, Sylvain Gelly, Jakob Uszkoreit, and Neil Houlsby.
\newblock {An Image is Worth 16x16 Words: Transformers for Image Recognition at
  Scale}.
\newblock In \emph{ICLR}, 2021.

\bibitem[Elfwing et~al.(2018)Elfwing, Uchibe, and Doya]{silu}
Stefan Elfwing, Eiji Uchibe, and Kenji Doya.
\newblock {Sigmoid-weighted linear units for neural network function
  approximation in reinforcement learning}.
\newblock \emph{Neural networks}, 107:\penalty0 3--11, 2018.

\bibitem[Etcheverry et~al.(2026)Etcheverry, Miotti, Sirbu, Sch{\"u}rholt,
  Drozdova, Ghosh, y~Arcas, Manyika, Richards, and
  Niklasson]{etcheverry2026nca}
Mayalen Etcheverry, Pietro Miotti, Aidan Sirbu, Konstantin Sch{\"u}rholt,
  Mariia Drozdova, Arna Ghosh, Blaise~Ag{\"u}era y Arcas, James Manyika, Blake
  Richards, and Eyvind Niklasson.
\newblock {Reasoning with Neural Cellular Automata}.
\newblock arXiv preprint arXiv:2609.36126, 2026.

\bibitem[Gao et~al.(2026)Gao, Sch{\"o}lkopf, and Geiger]{gao2026loopowm}
Gege Gao, Bernhard Sch{\"o}lkopf, and Andreas Geiger.
\newblock {Slots, Transitions, Loops: Learning Composable World Models for
  ARC}.
\newblock arXiv preprint arXiv:2606.12316, 2026.

\bibitem[Geiping et~al.(2025)Geiping, McLeish, Jain, Kirchenbauer, Singh,
  Bartoldson, Kailkhura, Bhatele, and Goldstein]{geiping2025huginn}
Jonas Geiping, Sean McLeish, Neel Jain, John Kirchenbauer, Siddharth Singh,
  Brian Bartoldson, Bhavya Kailkhura, Abhinav Bhatele, and Tom Goldstein.
\newblock {Scaling up Test-Time Compute with Latent Reasoning: A Recurrent
  Depth Approach}.
\newblock In \emph{NeurIPS}, 2025.

\bibitem[Giannou et~al.(2023)Giannou, Rajput, Sohn, Lee, Lee, and
  Papailiopoulos]{giannou2023looped}
Angeliki Giannou, Shashank Rajput, Jy-Yong Sohn, Kangwook Lee, Jason~D. Lee,
  and Dimitris Papailiopoulos.
\newblock {Looped Transformers as Programmable Computers}.
\newblock In \emph{ICML}, 2023.

\bibitem[Graves(2016)]{graves2016act}
Alex Graves.
\newblock {Adaptive Computation Time for Recurrent Neural Networks}.
\newblock arXiv preprint arXiv:1603.08983, 2016.

\bibitem[He et~al.(2016)He, Zhang, Ren, and Sun]{resnet}
Kaiming He, Xiangyu Zhang, Shaoqing Ren, and Jian Sun.
\newblock {Deep Residual Learning for Image Recognition}.
\newblock In \emph{CVPR}, 2016.

\bibitem[Hodel(2024)]{rearc}
Michael Hodel.
\newblock {Addressing the Abstraction and Reasoning Corpus via Procedural
  Example Generation}.
\newblock \emph{arXiv preprint arXiv:2404.07353}, 2024.

\bibitem[Hu et~al.(2026)Hu, Cy, Qiu, Ding, Wang, Zhu, Andreas, and
  He]{hu2026varc}
Keya Hu, Ali Cy, Linlu Qiu, Xiaoman~Delores Ding, Runqian Wang, Yeyin~Eva Zhu,
  Jacob Andreas, and Kaiming He.
\newblock {ARC Is a Vision Problem!}
\newblock In \emph{CVPR}, 2026.

\bibitem[Huang et~al.(2018)Huang, Chen, Li, Wu, van~der Maaten, and
  Weinberger]{huang2018msdnet}
Gao Huang, Danlu Chen, Tianhong Li, Felix Wu, Laurens van~der Maaten, and
  Kilian~Q. Weinberger.
\newblock {Multi-Scale Dense Networks for Resource Efficient Image
  Classification}.
\newblock In \emph{ICLR}, 2018.

\bibitem[Jeddi et~al.(2026)Jeddi, Ciccone, and Taati]{jeddi2026loopformer}
Ahmadreza Jeddi, Marco Ciccone, and Babak Taati.
\newblock {LoopFormer: Elastic-Depth Looped Transformers for Latent Reasoning
  via Shortcut Modulation}.
\newblock In \emph{ICLR}, 2026.

\bibitem[Jolicoeur-Martineau(2025)]{jolicoeur2025trm}
Alexia Jolicoeur-Martineau.
\newblock {Less is More: Recursive Reasoning with Tiny Networks}.
\newblock arXiv preprint arXiv:2510.04871, 2025.

\bibitem[Kornblith et~al.(2019)Kornblith, Norouzi, Lee, and Hinton]{cka}
Simon Kornblith, Mohammad Norouzi, Honglak Lee, and Geoffrey Hinton.
\newblock {Similarity of neural network representations revisited}.
\newblock In \emph{ICML}, 2019.

\bibitem[Kuo et~al.(2026)Kuo, Chayti, Reizinger, Brendel, and
  Jaggi]{kuo2026stochastic}
Hsun-Yu Kuo, El~Mahdi Chayti, Patrik Reizinger, Wieland Brendel, and Martin
  Jaggi.
\newblock {Stabilizing Extrapolation in Looped Transformers via Learned
  Stochastic Stopping}.
\newblock \emph{arXiv preprint arXiv:2606.29983}, 2026.

\bibitem[LeCun et~al.(1998)LeCun, Bottou, Bengio, and Haffner]{mnist}
Yann LeCun, L{\'e}on Bottou, Yoshua Bengio, and Patrick Haffner.
\newblock {Gradient-based learning applied to document recognition}.
\newblock \emph{Proceedings of the IEEE}, 86\penalty0 (11):\penalty0
  2278--2324, 1998.

\bibitem[Lin et~al.(2017)Lin, Doll{\'a}r, Girshick, He, Hariharan, and
  Belongie]{fpn}
Tsung-Yi Lin, Piotr Doll{\'a}r, Ross Girshick, Kaiming He, Bharath Hariharan,
  and Serge Belongie.
\newblock {Feature Pyramid Networks for Object Detection}.
\newblock In \emph{CVPR}, 2017.

\bibitem[Liu et~al.(2026)Liu, Ma, Xie, and Hua]{liu2026tracevit}
Binnan Liu, Yechi Ma, Tian Xie, and Wei Hua.
\newblock {TraceViT: Grounded Trace Supervision for Visual Abstract Reasoning}.
\newblock arXiv preprint arXiv:2607.29586, 2026.

\bibitem[Loshchilov and Hutter(2019)]{adamw}
Ilya Loshchilov and Frank Hutter.
\newblock {Decoupled Weight Decay Regularization}.
\newblock In \emph{ICLR}, 2019.

\bibitem[Miyato et~al.(2025)Miyato, L{\"o}we, Geiger, and Welling]{akon}
Takeru Miyato, Sindy L{\"o}we, Andreas Geiger, and Max Welling.
\newblock {Artificial Kuramoto Oscillatory Neurons}.
\newblock In \emph{ICLR}, 2025.

\bibitem[Nie et~al.(2020)Nie, Yu, Mao, Patel, Zhu, and
  Anandkumar]{nie2020bongardlogo}
Weili Nie, Zhiding Yu, Lei Mao, Ankit~B. Patel, Yuke Zhu, and Anima Anandkumar.
\newblock {Bongard-LOGO: A New Benchmark for Human-Level Concept Learning and
  Reasoning}.
\newblock In \emph{NeurIPS}, 2020.

\bibitem[Palm et~al.(2018)Palm, Paquet, and Winther]{rrn}
Rasmus Palm, Ulrich Paquet, and Ole Winther.
\newblock {Recurrent relational networks}.
\newblock In \emph{NeurIPS}, 2018.

\bibitem[Perez et~al.(2018)Perez, Strub, de~Vries, Dumoulin, and
  Courville]{perez2018film}
Ethan Perez, Florian Strub, Harm de Vries, Vincent Dumoulin, and Aaron
  Courville.
\newblock {FiLM: Visual Reasoning with a General Conditioning Layer}.
\newblock In \emph{AAAI}, 2018.

\bibitem[Schwarzschild et~al.(2021)Schwarzschild, Borgnia, Gupta, Huang,
  Vishkin, Goldblum, and Goldstein]{schwarzschild2021algorithm}
Avi Schwarzschild, Eitan Borgnia, Arjun Gupta, Furong Huang, Uzi Vishkin, Micah
  Goldblum, and Tom Goldstein.
\newblock {Can You Learn an Algorithm? Generalizing from Easy to Hard Problems
  with Recurrent Networks}.
\newblock In \emph{NeurIPS}, 2021.

\bibitem[Shu et~al.(2026)Shu, Qiu, Zhu, Chen, Liu, and Yang]{shu2026loopvit}
Wen-Jie Shu, Xuerui Qiu, Rui-Jie Zhu, Harold~Haodong Chen, Yexin Liu, and Harry
  Yang.
\newblock {{L}oop{V}i{T}: Scaling Visual {ARC} with Looped Transformers}.
\newblock In \emph{Findings of the {A}ssociation for {C}omputational
  {L}inguistics: {EMNLP} 2026}, 2026.

\bibitem[Tian et~al.(2020)Tian, Krishnan, and Isola]{cmc}
YongLong Tian, Dilip Krishnan, and Phillip Isola.
\newblock {Contrastive Multiview Coding}.
\newblock In \emph{ECCV}, 2020.

\bibitem[Tian et~al.(2023)Tian, Xie, Wang, Wei, Zhang, Jiao, Wang, Tian, and
  Ye]{itpn}
Yunjie Tian, Lingxi Xie, Zhaozhi Wang, Longhui Wei, Xiaopeng Zhang, Jianbin
  Jiao, Yaowei Wang, Qi Tian, and Qixiang Ye.
\newblock {Integrally Pre-Trained Transformer Pyramid Networks}.
\newblock In \emph{CVPR}, 2023.

\bibitem[Wang et~al.(2025)Wang, Li, Sun, Chen, Liu, Wu, Lu, Song, and
  Yadkori]{wang2025hrm}
Guan Wang, Jin Li, Yuhao Sun, Xing Chen, Changling Liu, Yue Wu, Meng Lu, Sen
  Song, and Yasin~Abbasi Yadkori.
\newblock {Hierarchical Reasoning Model}.
\newblock arXiv preprint arXiv:2506.21734, 2025.

\bibitem[Wang et~al.(2019)Wang, Donti, Wilder, and Kolter]{satnet}
Po-Wei Wang, Priya Donti, Bryan Wilder, and Zico Kolter.
\newblock {SATNet: Bridging deep learning and logical reasoning using a
  differentiable satisfiability solver}.
\newblock In \emph{ICML}, 2019.

\bibitem[Xie et~al.(2026)Xie, Wei, Cao, Zhao, Deng, Li, Dai, Gao, Xu, Yu, Zhao,
  Zhou, Xu, Zhang, Zeng, Hu, Wang, Yuan, Wang, and Liang]{mhc}
Zhenda Xie, Yixuan Wei, Huanqi Cao, Chenggang Zhao, Chengqi Deng, Jiashi Li,
  Damai Dai, Huazuo Gao, Mingyu Xu, Kuai Yu, Liang Zhao, Shangyan Zhou, Zhean
  Xu, Zhengyan Zhang, Wangding Zeng, Shengding Hu, Yuqing Wang, Jingyang Yuan,
  Lean Wang, and Wenfeng Liang.
\newblock {mHC: Manifold-Constrained Hyper-Connections}.
\newblock In \emph{ICML}, 2026.

\bibitem[Yun et~al.(2019)Yun, Han, Oh, Chun, Choe, and Yoo]{cutmix}
Sangdoo Yun, Dongyoon Han, Seong~Joon Oh, Sanghyuk Chun, Junsuk Choe, and
  Youngjoon Yoo.
\newblock {Cutmix: Regularization strategy to train strong classifiers with
  localizable features}.
\newblock In \emph{ICCV}, 2019.

\bibitem[Zhang and Sennrich(2019)]{rmsnorm}
Biao Zhang and Rico Sennrich.
\newblock {Root Mean Square Layer Normalization}.
\newblock In \emph{NeurIPS}, 2019.

\bibitem[Zhang et~al.(2019)Zhang, Gao, Jia, Zhu, and Zhu]{zhang2019raven}
Chi Zhang, Feng Gao, Baoxiong Jia, Yixin Zhu, and Song-Chun Zhu.
\newblock {RAVEN: A Dataset for Relational and Analogical Visual REasoNing}.
\newblock In \emph{CVPR}, 2019.

\bibitem[Zhang et~al.(2018)Zhang, Cisse, Dauphin, and Lopez-Paz]{mixup}
Hongyi Zhang, Moustapha Cisse, Yann~N. Dauphin, and David Lopez-Paz.
\newblock {mixup: Beyond Empirical Risk Minimization}.
\newblock In \emph{ICLR}, 2018.

\bibitem[Zhong et~al.(2020)Zhong, Zheng, Kang, Li, and Yang]{randomerase}
Zhun Zhong, Liang Zheng, Guoliang Kang, Shaozi Li, and Yi Yang.
\newblock {Random Erasing Data Augmentation}.
\newblock In \emph{AAAI}, 2020.

\bibitem[Zhou et~al.(2026)Zhou, Zhang, Huang, Li, Chen, and
  Wang]{zhou2026retain}
Peilong Zhou, Hongtao Zhang, Yulong Huang, Cangyuan Li, Mingyu Chen, and Ying
  Wang.
\newblock {Learning to Compute Is Not Learning to Retain or Reuse in Looped
  Transformers}.
\newblock \emph{arXiv preprint arXiv:2610.19768}, 2026.

\bibitem[Zhu et~al.(2025)Zhu, Wang, Hua, Zhang, Li, Que, Wei, Wen, Yin, Xing,
  Li, Shi, Ma, Li, Kergan, Smith, Qu, Hui, Wu, Min, Huang, Zhou, Ye, Liu, Yang,
  Shi, Lin, Zhao, Cai, Zhang, Huang, Bengio, and Eshraghian]{zhu2025ouro}
Rui-Jie Zhu, Zixuan Wang, Kai Hua, Tianyu Zhang, Ziniu Li, Haoran Que, Boyi
  Wei, Zixin Wen, Fan Yin, He Xing, Lu Li, Jiajun Shi, Kaijing Ma, Shanda Li,
  Taylor Kergan, Andrew Smith, Xingwei Qu, Mude Hui, Bohong Wu, Qiyang Min,
  Hongzhi Huang, Xun Zhou, Wei Ye, Jiaheng Liu, Jian Yang, Yunfeng Shi,
  Chenghua Lin, Enduo Zhao, Tianle Cai, Ge Zhang, Wenhao Huang, Yoshua Bengio,
  and Jason Eshraghian.
\newblock {Scaling Latent Reasoning via Looped Language Models}.
\newblock arXiv preprint arXiv:2510.25741, 2025.

\end{thebibliography}
}

\clearpage

\onecolumn

\appendix
\section{Implementation Detail}
\label{app:experimental-settings}

\subsection{Datasets and Evaluation}

We evaluate on the public evaluation sets of ARC-AGI-1 and ARC-AGI-2, containing 400 and 120 tasks, respectively. Offline pretraining uses the 400 ARC-AGI-1 training tasks together with synthetic examples from RE-ARC. For ARC-AGI-2, we use the ARC-AGI-1-pretrained models and perform task-specific test-time training (TTT) on the provided demonstration pairs. During TTT, all model parameters are fine-tuned using augmented demonstrations. Predictions from two independent adaptation runs, each with ten inference augmentations, are transformed back to the original coordinates and aggregated by voting over complete output grids. For pass@2, we count a query as correct if either of the two highest-voted distinct predictions matches the target grid in both dimensions and all cell values. Query scores are averaged within each task and then across tasks.

\subsection{Model Configuration}

Table~\ref{tab:arc-model-configs} summarizes the three LoopSight scales used for ARC-AGI. All variants use a $64\times64$ canvas, $2\times2$ patches, eight attention heads, and a nominal feed-forward hidden dimension of 512. Core parameters are shared across loop iterations.

\begin{table}[htbp]
    \centering
    \small
    \caption{LoopSight model configurations and scale-specific training settings for ARC-AGI.}
    \label{tab:arc-model-configs}
    \begin{tabular*}{0.9\linewidth}{@{\extracolsep{\fill}}lccc@{}}
        \toprule
        Setting & Small & Medium & Large \\
        \midrule
        Embedding dimension $D$ & 256 & 384 & 512 \\
        Shared Transformer blocks & 6 & 8 & 10 \\
        Write-gate bottleneck width $r$ & 16 & 24 & 32 \\
        Default loop count for TTT and inference & 12 & 6 & 6 \\
        \midrule
        Pretraining loop range $[N_{\min},N_{\max}]$ & 6--18 & 3--9 & 3--9 \\
        Gaussian center $\mu$ & 12 & 6 & 6 \\
        Gaussian scale $\tau$ & 3 & 1.5 & 1.5 \\
        Peak pretraining learning rate & $3\times10^{-4}$ & $3\times10^{-4}$ & $2.5\times10^{-4}$ \\
        \bottomrule
    \end{tabular*}
\end{table}

\subsection{Gate and Encoding Setting}

The write gate uses a bottleneck width of $r=D/16$, where $D$ is the embedding dimension. We set the gate logit bound to $b=\ln 16$ and zero-initialize the final gate projection and its bias, yielding initial write coefficients $\alpha_n=1/n$. Asymptotic loop encoding uses eight learned geometric channels, complemented by three cosine features for the within-loop block position. Each block uses a modulation MLP with hidden dimension 64. Token-wise RMSNorm~\citep{rmsnorm} is applied after each complete loop.

\subsection{Training Protocol}

During offline pretraining, we sample one loop count per optimizer update from the truncated discrete Gaussian in Eq.~\eqref{eq:depth-sampling}, sharing it across distributed workers and accumulated microbatches. We supervise only the final prediction at the sampled depth using pixel-wise cross-entropy, with full backpropagation through all iterations. Models are trained for 100 epochs using AdamW~\citep{adamw}, a global batch size of 256, and cosine learning-rate decay with 10 warmup epochs. For each model scale, we use the same peak pretraining learning rate as in the corresponding LoopViT training configuration (Table~\ref{tab:arc-model-configs}). We use zero weight decay, dropout 0.1, and gradient clipping with a maximum norm of 1.

We maintain an exponential moving average (EMA) of the model parameters during offline pretraining with a decay of 0.9999. Each TTT attempt is initialized from the EMA checkpoint exported at the final pretraining epoch. During TTT, we maintain an independent EMA for each task and adaptation attempt with a decay of 0.998. For EMA-based evaluation, predictions are generated from these task-specific averaged weights and aggregated by voting over complete output grids, as described above. With both offline and TTT EMA disabled, LoopSight-Small still achieves 66.6\% Pass@2 on ARC-AGI-1 at 12 loops, compared with 60.8\% for LoopViT-Small under the same loop budget.

\subsection{Progressive Ablation Setting}

All variants use three shared Transformer blocks, a hidden dimension of 192,
three attention heads, and a feed-forward dimension of 512.
We pretrain for 50 epochs on 192 ARC-AGI-1 training tasks augmented with
RE-ARC, using AdamW with a peak learning rate of $7\times10^{-4}$.
The first three variants use six loops during pretraining. The final variant
samples 3--9 loops from a truncated discrete Gaussian ($\mu=6$, $\tau=1.5$).
On 192 evaluation tasks, we run two 50-epoch TTT attempts per task at six
loops, with a peak learning rate of $4.5\times10^{-4}$.

\subsection{Visual Sudoku}

Visual Sudoku requires completing a $9\times9$ board rendered with handwritten MNIST digits~\citep{mnist}. We use $252\times252$ grayscale inputs and $14\times14$ patches. Training uses 9,000 easy SATNet puzzles~\citep{satnet}, while the Easy split contains 1,000 held-out puzzles from the same puzzle distribution. For OOD evaluation, the Hard split contains 1,000 puzzles sampled without replacement from the 18,000-puzzle RRN~\citep{rrn} test set using random seed 0~\citep{akon}. This subset is fixed across all models, checkpoints, and inference depths. Training and evaluation use disjoint MNIST glyph pools, with fresh glyphs sampled during training and deterministic rendering during evaluation.

LoopViT and LoopSight both use hidden dimension 256, eight shared Transformer blocks, and eight attention heads. As this task does not require test-time training on new data, we train LoopSight with a fixed budget of 12 loops and retain the same budget at evaluation; LoopViT uses the same loop count. Both models are trained from scratch for 200 epochs using AdamW, a batch size of 64, a peak learning rate of $2\times10^{-4}$, and weight decay of $10^{-4}$. We use 5\% linear warmup followed by cosine decay to zero, dropout of 0.3, gradient clipping at 1.0, and BF16 mixed precision. Training minimizes pixel MSE on the final reconstructed image, and the primary EMA decay is 0.9995. Predictions are decoded using fixed MNIST class-mean prototypes, and whole-board accuracy requires all 81 cells to be correct. Easy also serves as the validation set for checkpoint selection, while Hard is reserved for evaluation.

\subsection{ImageNet-100 Classification}

We evaluate on the CMC selection of ImageNet-100~\citep{cmc}, comprising 126,689 training images and 5,000 validation images from ImageNet-1K~\citep{imagenet}. All models use $224\times224$ RGB inputs and $16\times16$ patches. The Small versions of LoopViT and LoopSight use hidden dimension 256, six shared Transformer blocks, and eight attention heads. We replace their pixel-reconstruction heads with pooled classifiers: LoopViT pools the final recurrent patch states, whereas LoopSight pools the final GPM states. The baseline ViT-Small/16 uses hidden dimension 384 and 12 Transformer blocks. As this task does not require test-time training on new data, we train LoopSight with a fixed budget of six loops and retain the same budget at evaluation; LoopViT uses the same loop count.

All models are trained from scratch for 100 epochs using AdamW with a peak learning rate of $5\times10^{-4}$, weight decay of 0.05, and an effective batch size of 1,024. The recurrent models accumulate four microbatches of 256 images. We use five epochs of linear warmup followed by cosine decay to $10^{-6}$, dropout of 0.2, gradient clipping at 1.0, EMA decay of 0.9999, and BF16 mixed precision. All models share the same augmentation recipe, including random cropping and flipping, RandAugment~\citep{randaug}, Mixup~\citep{mixup}, CutMix~\citep{cutmix}, and random erasing~\citep{randomerase}. We evaluate raw and EMA weights every 10 epochs using Top-1 and Top-5 accuracy.

\begin{figure*}[h]
    \centering
    \begin{minipage}[c]{0.48\textwidth}
        \setlength{\parindent}{0pt}
        \section{Test-Time Scaling}
        
        We study how the number of recurrent loops affects test-time training (TTT) performance on ARC-AGI-1. Both models are initialized from their respective offline checkpoints and adapted to each evaluation task. We evaluate loop counts $K \in \{4, 8, 12, 20, 28\}$ and report Pass@2 over all 400 evaluation tasks, combining predictions from two adaptation attempts.
        
        As shown in Figure~\ref{fig:curve-ttt}, LoopSight-Small improves from 57.9\% at 4 loops to 66.1\% at 8 loops, and further increases to 68.0\% at 28 loops. LoopViT-Small improves from 54.6\% at 4 loops to a peak of 61.5\% at 20 loops, followed by a slight decline to 60.9\% at 28 loops.
        \medskip
    \end{minipage}
    \hfill
    \begin{minipage}[c]{0.45\textwidth}
        \centering
        \includegraphics[width=\linewidth]{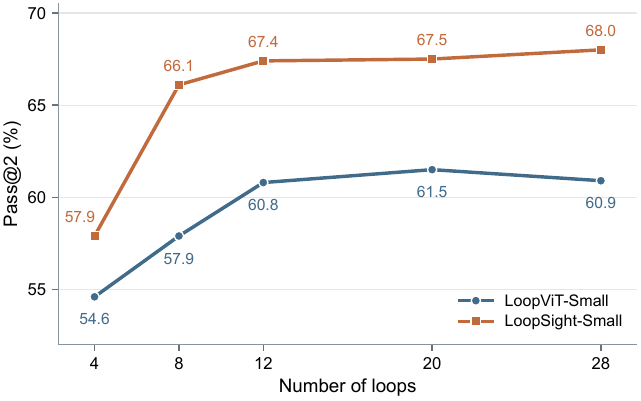}
        \caption{Pass@2 on ARC-AGI-1 versus the number of inference loops after task-specific TTT.}
        \label{fig:curve-ttt}
    \end{minipage}
\end{figure*}

LoopSight-Small maintains an advantage across the entire evaluated range. Its margin over LoopViT-Small increases from 3.3 percentage points at 4 loops to 7.1 percentage points at 28 loops. The continued improvement at larger loop counts suggests that LoopSight can make effective use of additional iterative computation after task-specific adaptation.

\section{Qualitative Visualization}

\subsection{ARC-AGI-1}
\label{sec:qualitative-arc-agi-1}

\begin{figure}[t]
    \centering
    \includegraphics[width=0.9\linewidth]{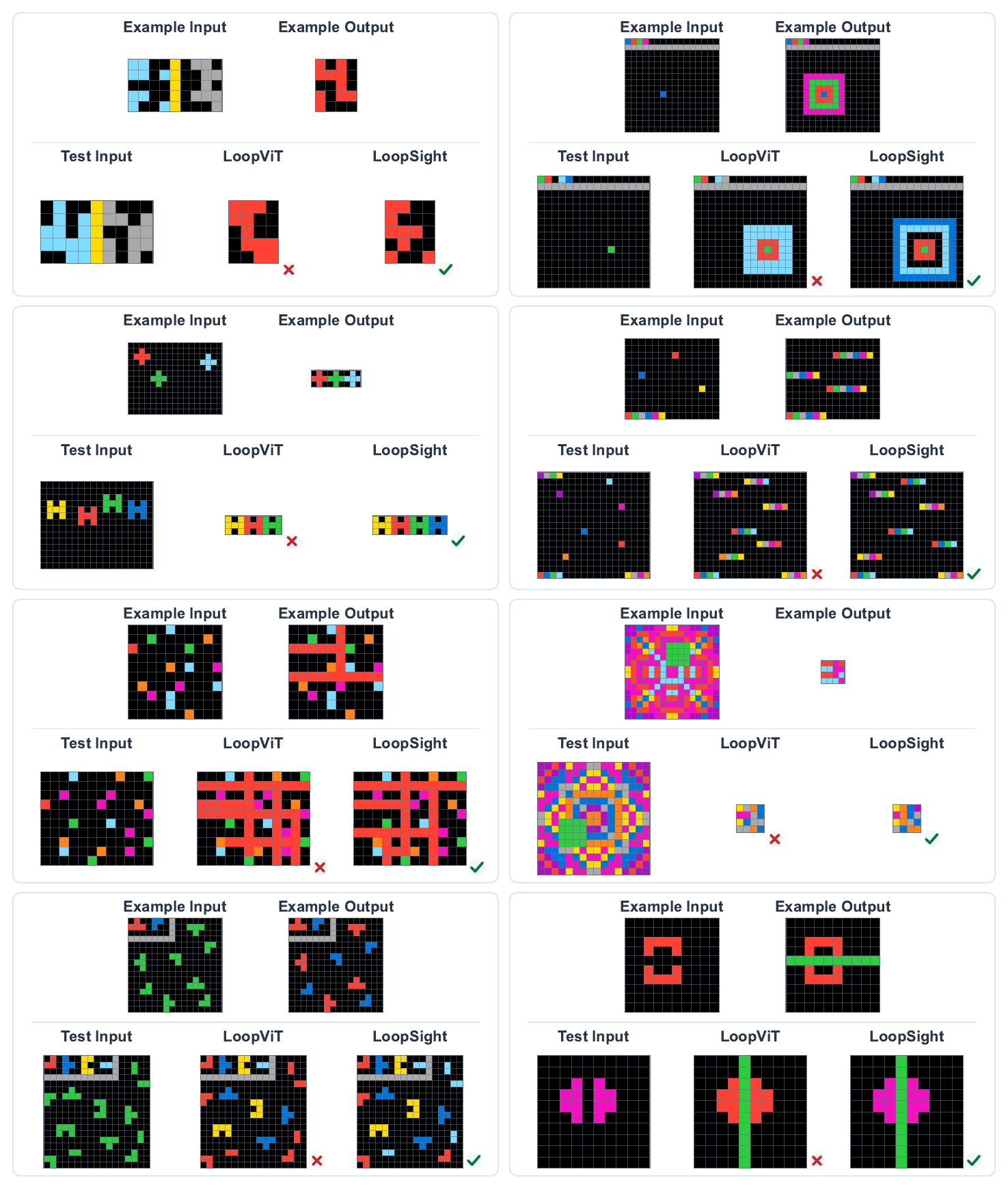}
    \caption{Selected qualitative comparisons on ARC-AGI-1. Each group shows the test input, the highest-voted LoopViT-Small
    prediction, and the exact match among the two
    highest-voted LoopSight-Small predictions.}
    \label{fig:arc-agi-1-qualitative}
\end{figure}

Figure~\ref{fig:arc-agi-1-qualitative} shows eight examples from the
ARC-AGI-1 public evaluation set. In each example, both of
LoopViT-Small's top two predictions are incorrect, and one of
LoopSight-Small's top two predictions exactly matches the target.
Both models use 12 loops for test-time training and inference.
Each group shows the test input, LoopViT-Small's highest-voted
prediction, and LoopSight-Small's matching prediction.

LoopViT-Small often recovers part of the target structure in these
examples. In one case, it places three motifs in the correct order
but omits the final blue motif, producing a grid that is three columns
too narrow. In another, it captures the center of a nested-square
pattern but misses the outer blue border. In a third case, it preserves
the shape of an object and the vertical line through it, but colors
the object red instead of magenta. These examples illustrate how errors
in grid dimensions, object structure, or color assignment can prevent
an otherwise close prediction from matching the target exactly.

\subsection{ARC-AGI-2}
\label{sec:qualitative-arc-agi-2}

\begin{figure}[t]
    \centering
    \includegraphics[width=0.9\linewidth]{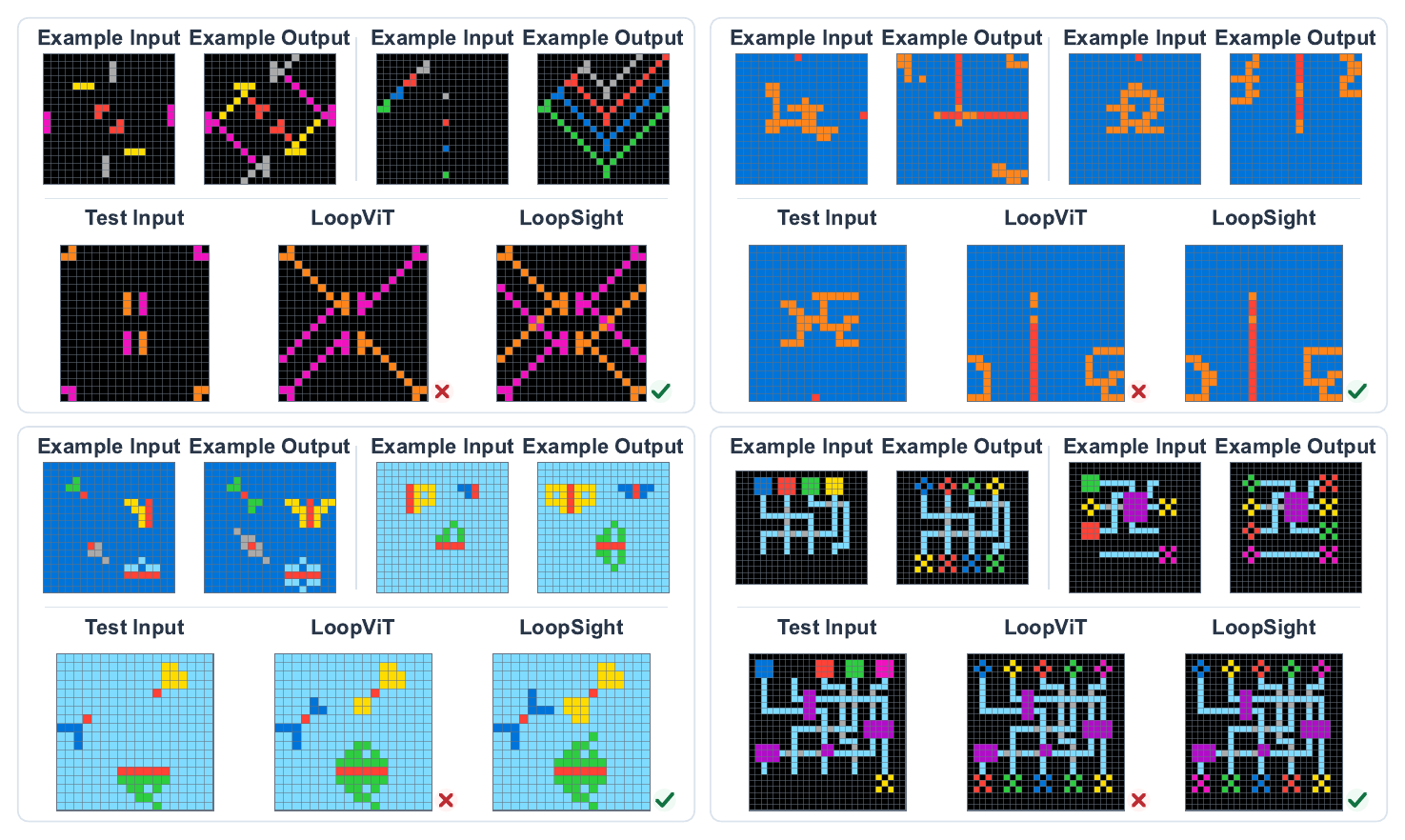}
    \caption{Selected qualitative comparisons on ARC-AGI-2.
    Each group shows two demonstration pairs above the test input,
    the highest-voted LoopViT-Small prediction, and the exact match
    among the two highest-voted LoopSight-Small predictions.}
    \label{fig:arc-agi-2-qualitative}
\end{figure}

Figure~\ref{fig:arc-agi-2-qualitative} shows four examples from the
ARC-AGI-2 public evaluation set. In each example, both
LoopViT-Small's top two predictions are incorrect, and one of
LoopSight-Small's top two predictions exactly matches the target.
Both models use 12 loops for test-time training and inference.
Each group includes two input--output demonstration pairs from
the same task.

LoopViT-Small recovers part of the target structure in four
examples. In the upper-left example, it draws the main diagonals
but omits segments that should reach the grid boundaries.
In the upper-right example, it leaves a gap in a vertical red line.
In the lower-left example, it omits cells in the blue and yellow
patterns. In the lower-right example, it places the bottom motifs
in the correct positions but assigns the wrong colors to two of them.
These examples show how errors in line continuity, pattern completion,
and color assignment can prevent a close prediction from
matching the target exactly.

\subsection{Failure Cases}
\label{sec:qualitative-failure-cases}

\begin{figure}[htbp]
    \centering
    \includegraphics[width=0.9\linewidth]{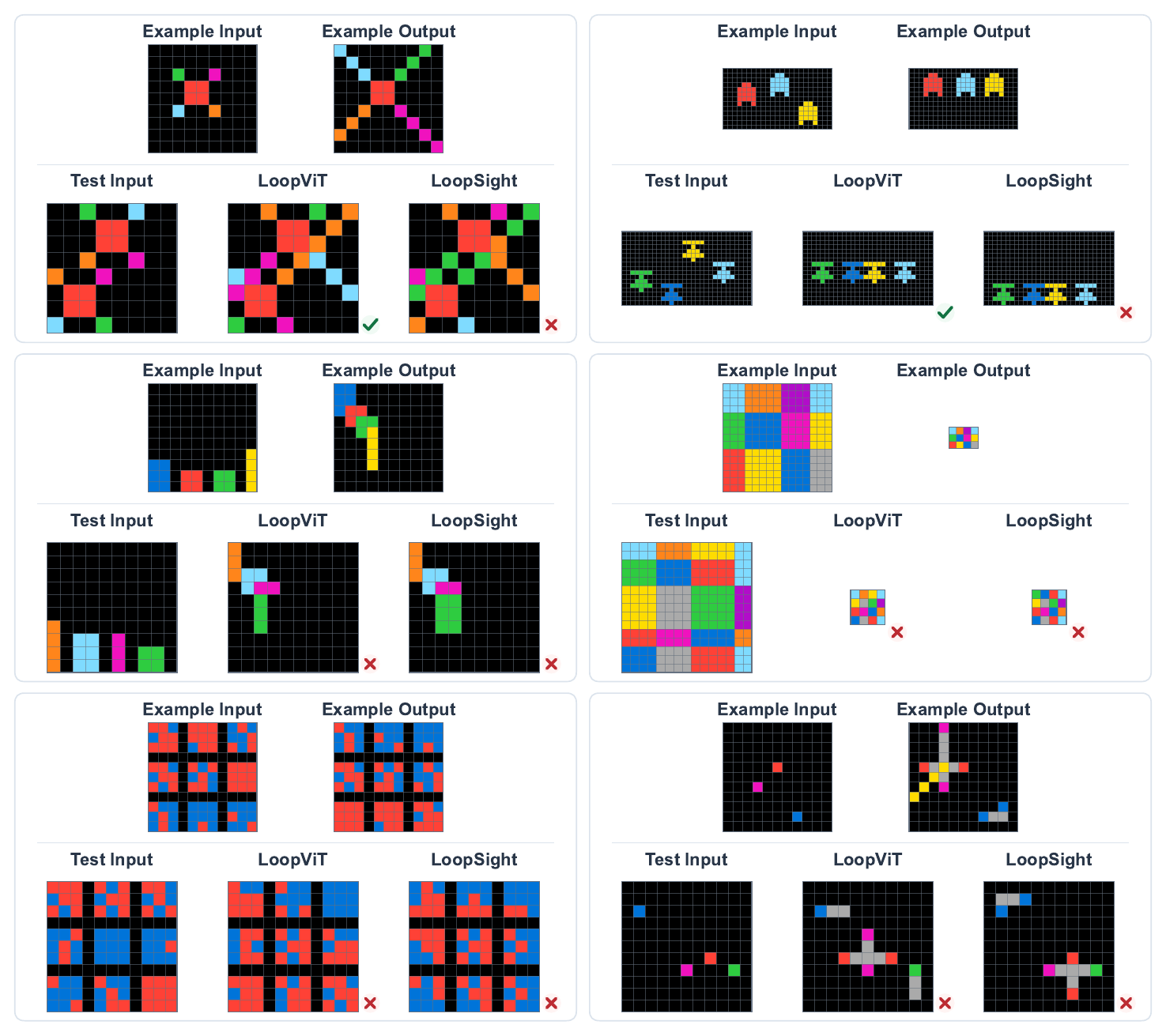}
    \caption{LoopSight-Small failure cases.}
    \label{fig:arc-agi-1-failure-cases}
\end{figure}

Figure~\ref{fig:arc-agi-1-failure-cases} shows six examples from the
ARC-AGI-1 evaluation set in which neither of LoopSight-Small's
top two predictions matches the target.

In one example, LoopSight-Small preserves the shapes and horizontal
order of four objects but places the entire row five cells too low.
In another, both models compress a grid of colored blocks into a
smaller grid, producing $4 \times 4$ outputs where the target is
$5 \times 4$. These examples suggest that exact spatial alignment and
output dimensions remain challenging even when the prediction captures
part of the demonstrated transformation. Gated Prediction Memory
aggregates features across loops, while exact-grid success still
depends on inferring the spatial constraints that determine the final
output.

\subsection{Visual Sudoku}
\label{sec:qualitative-visual-sudoku}

\begin{figure*}[t]
    \centering
    \includegraphics[width=1.\textwidth]{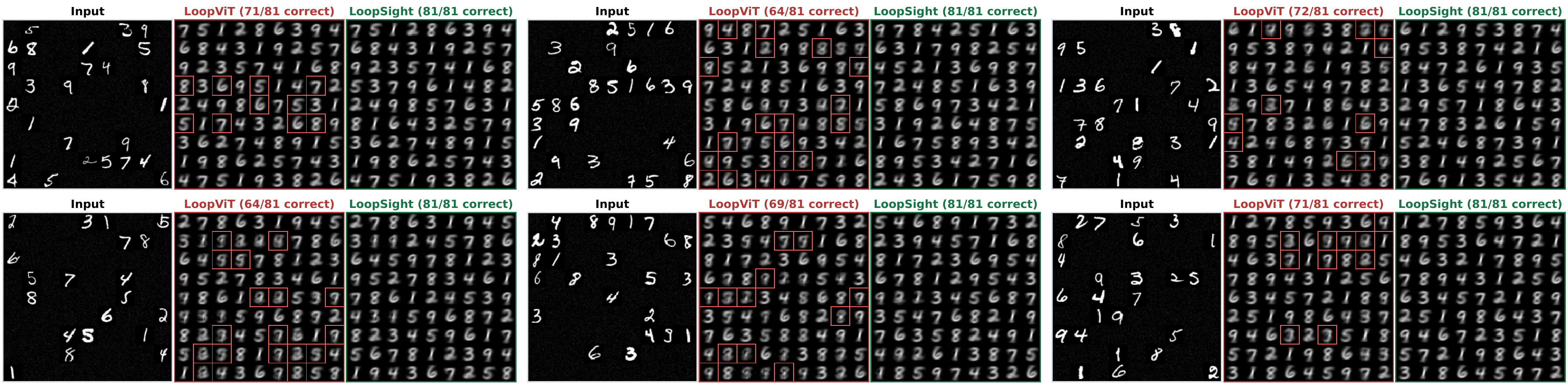}
    \caption{Selected qualitative comparisons on Visual Sudoku.
    Each group shows the input, LoopViT output, and LoopSight output.
    Labels report correct cells out of 81; red boxes mark incorrect predictions.}
    \label{fig:visual-sudoku-qualitative}
\end{figure*}

Figure~\ref{fig:visual-sudoku-qualitative} shows six illustrative
examples from the Hard split, selected to contrast successful
LoopSight completions with unsuccessful LoopViT predictions.
Both models receive identical input images and use 12 inference
loops with one prediction per board. We use their final EMA
checkpoints after 200 epochs under the same training recipe.
The visualizations show directly generated grayscale images,
evaluated using the fixed MNIST-prototype decoder without
clue correction or solver-based refinement.

In these examples, both models generate recognizable digit shapes.
LoopViT's prediction errors occur primarily in initially blank
cells and extend across multiple rows and columns; one example
also contains an incorrectly predicted given clue. These patterns
illustrate the need to coordinate individual digit predictions
with constraints across the board. The selected comparisons
highlight how differences in completing a globally consistent
solution can remain visible even when both models produce
readable digit images.

\end{document}